\documentclass{article}
\usepackage{spconf}
\usepackage{epsfig}
\usepackage{romannum}
\usepackage{amsmath}
\usepackage{multirow}
\usepackage{multicol}
\usepackage{booktabs}
\usepackage{graphicx}
\usepackage{boldline}
\usepackage[table, dvipsnames]{xcolor}
\usepackage{indentfirst}
\usepackage{subfigure}
\usepackage[belowskip=0pt,aboveskip=5pt]{caption}
\usepackage{mathtools}
\usepackage{amssymb}
\usepackage{amsbsy}
\usepackage{stackengine}
\usepackage{hyperref}
\hypersetup{
	colorlinks = true, 
	urlcolor = red, 
	linkcolor = red, 
	citecolor = green}

\title{CRVOS: Clue Refining Network for Video Object Segmentation}

\name{
	Suhwan Cho\sthanks{E-mail: chosuhwan@yonsei.ac.kr}, MyeongAh Cho, Tae-young Chung, Heansung Lee, Sangyoun Lee\sthanks{Corresponding Author, E-mail: syleee@yonsei.ac.kr}}
\address{Yonsei University, Seoul, Korea}

\begin{document}
\maketitle
\begin{abstract}

The encoder-decoder based methods for semi-supervised video object segmentation (Semi-VOS) have received extensive attention due to their superior performances. However, most of them have complex intermediate networks which generate strong specifiers to be robust against challenging scenarios, and this is quite inefficient when dealing with relatively simple scenarios. To solve this problem, we propose a real-time network, Clue Refining Network for Video Object Segmentation (CRVOS), that does not have any intermediate network to efficiently deal with these scenarios. In this work, we propose a simple specifier, referred to as the Clue, which consists of the previous frame’s coarse mask and coordinates information. We also propose a novel refine module which shows the better performance compared with the general ones by using a deconvolution layer instead of a bilinear upsampling layer. Our proposed method shows the fastest speed among the existing methods with a competitive accuracy. On DAVIS 2016 validation set, our method achieves 63.5 fps and $\mathcal{J}\&\mathcal{F}$ score of 81.6\%. 

\end{abstract}

\section{Introduction}
\label{intro}
Semi-supervised video object segmentation (Semi-VOS) is a task to find the labels of every pixel in video with the given mask, i.e. the initial frame's mask. Semi-VOS can be divided into two categories, with and without online fine-tuning. The methods with online fine-tuning \cite{caelles2017one, voigtlaender2017online, perazzi2017learning, hu2017maskrnn} generally show better accuracy compared to the ones without since they can prepare the most suitable environment for the given condition. However, they usually show slower speed due to the high computational cost of the fine-tuning process. On the other hand, the methods without online fine-tuning \cite{marki2016bilateral, jampani2017video, cheng2018fast, yang2018efficient} are generally faster than with the ones, but the accuracy is not comparable to them.

\begin{figure}[t]
	\centering
	\includegraphics[width=1.0\linewidth]{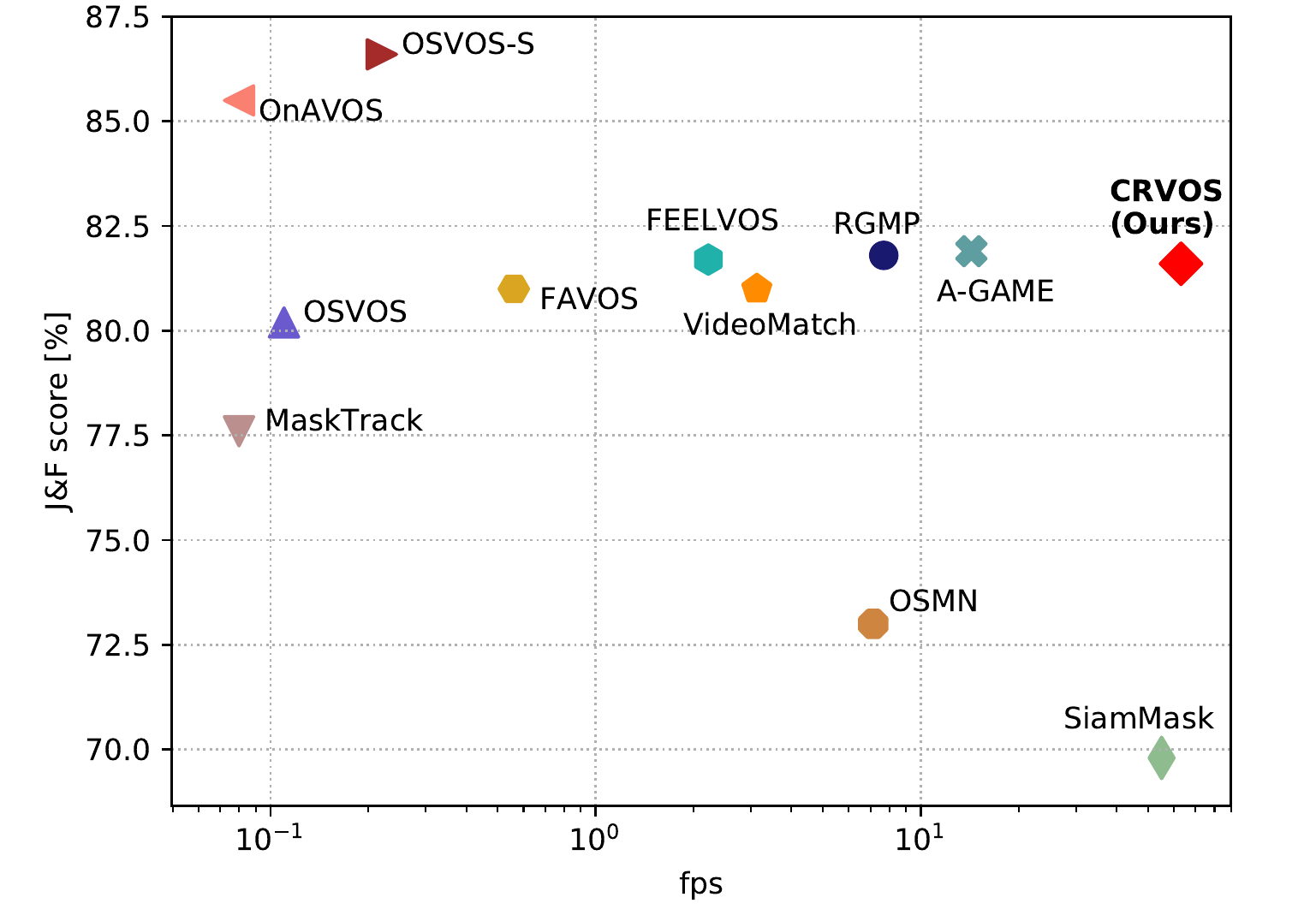}
	\caption{Comparison of quantitative methods on DAVIS 2016 validation set. We visualize $\mathcal{J}$\&$\mathcal{F}$ score with respect to fps. Note that fps axis is in the log scale.}
	\label{figure1}
\end{figure}

\begin{figure*}[t]
	\centering
	\includegraphics[width=1.0\linewidth]{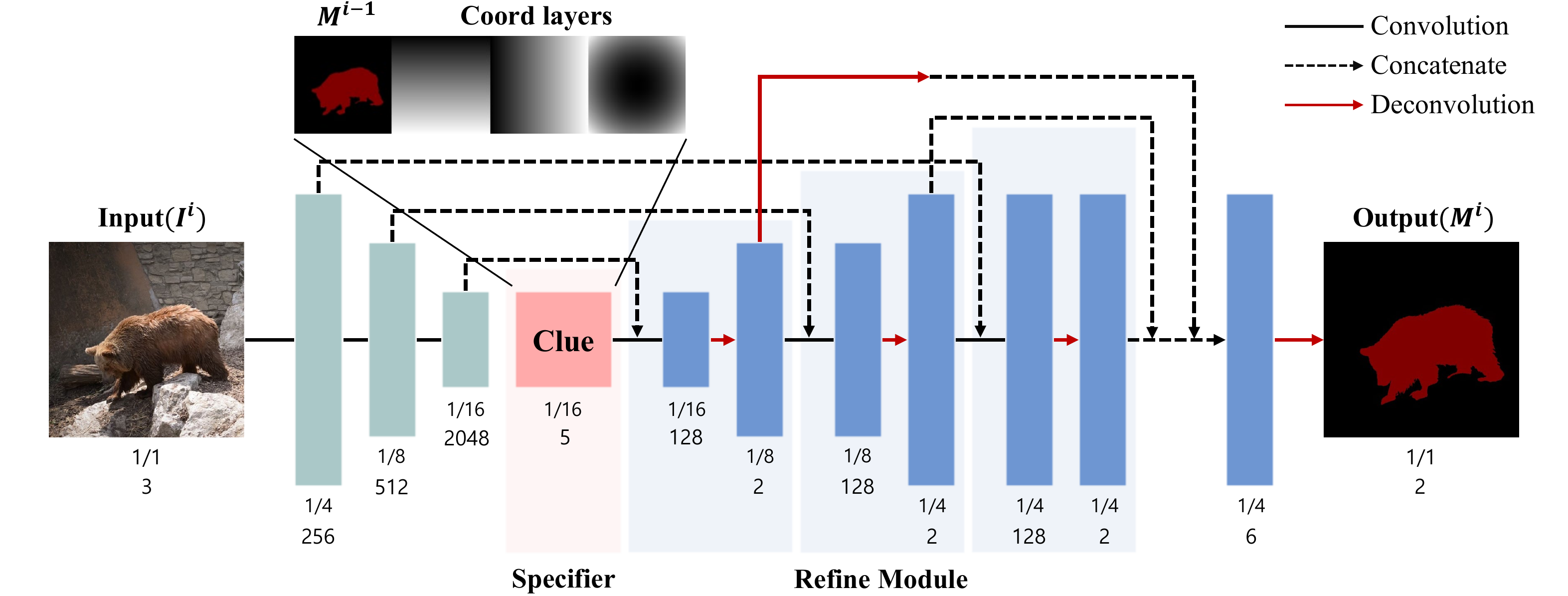}
	\caption{Network architecture of CRVOS, which consists of an encoder, a decoder, and the Clue. Decoder is composed of three refine modules which refine the Clue. $I^i$, $M^i$, and $M^{i-1}$ indicate the current frame's input image, the current frame's mask, and the previous frame's mask respectively.}
	\label{figure2}
\end{figure*}

Until recently, accuracy of the methods with and without online fine-tuning differed greatly due to the aforementioned reason. The accuracy gap between these methods was narrowed by RGMP \cite{wug2018fast}, which achieved satisfactory accuracy without online fine-tuning. Since it succeeded with encoder-decoder based architecture, many recent methods \cite{johnander2019generative, lin2019agss, zeng2019dmm, oh2019video, wang2019ranet} were designed based on encoder-decoder architecture and most of them showed competitive performances. It can be observed that they share the following algorithm. Encoder extracts the features from the current frame's input image. These features are connected to decoder with skip connections. The decoder predicts the current frame's mask with the features and some additional information.

Semi-VOS deals with an arbitrary target given in the initial frame, making it hard to specify the given target only with encoder and decoder. Therefore, the encoder-decoder based method has the intermediate network between its encoder and decoder to generate some additional information about the target. We call the additional information as a specifier since it is used for specifying the given target. For example, RGMP \cite{wug2018fast} uses Siamese encoder with Global Convolution Block, STM \cite{oh2019video} uses Space-Time Memory Networks, and RANet \cite{wang2019ranet} uses Ranking Attention Module as the intermediate network to generate the specifier. The problem arises because, these intermediate networks are generally complex and quite computationally expensive, but are only effective when dealing with challenging scenarios. In other words, the existing encoder-decoder based methods designed to have strong specifiers are not efficient in most of relatively simple scenarios.

In this work, we suggest a novel approach to Semi-VOS to effectively deal with the simple scenarios. Our method runs in real-time while maintaining a competitive accuracy. In Fig. \ref{figure1}, we compare quantitative methods on DAVIS 2016 \cite{perazzi2016benchmark} validation set. Our contributions can be summarized as follows.

\begin{itemize}
	\setlength{\itemsep}{0.5em}
	\setlength{\parskip}{0pt}
	\item We suggest a novel way to generate a simple yet effective specifier. Our specifier, the Clue, enables our network to be competitive with real-time performance. 
	\item We propose a novel refine module for Semi-VOS. It is more effective than general ones, since it uses a deconvolution layer instead of a bilinear upsampling layer. 
\end{itemize}

\section{Method}\label{method}
\subsection{Overview} 
The goal of Semi-VOS is to segment the given target in video, when the initial frame's mask is given. When the current frame's input image is received, the network should infer the current frame's mask with the accumulated information, e.g. the initial frame's input image, the previous frame's input image, and the previous frame's mask. In Fig. \ref{figure2}, we show network architecture of CRVOS, which consists of an encoder, a decoder, and the Clue. The encoder is based on ResNet-50 \cite{he2016deep}, pre-trained on ImageNet \cite{deng2009imagenet}. It extracts the features from the current frame's input image and these features are connected to refine modules with skip connections. The size of the last features is 1/16 of the input image size. The Clue is composed of the previous frame's coarse mask with coordinates information, and also has the size of 1/16 of the input image. The decoder predicts the current frame's mask by refining the Clue with three refine modules.

\subsection{Specifier}
Most of the current encoder-decoder based methods have complex intermediate networks to generate strong specifiers. However as mentioned earlier, these methods are not efficient in relatively simple scenarios. Therefore, to efficiently deal with these scenarios, we focus on finding a simple specifier that does not need any intermediate network, but still can specify the given target. We assume that the previous frame's coarse mask can be the solution. Since positional changes of the target between adjacent frames are generally small, the previous frame's mask has enough positional information about the current frame's target, and more so at the coarse level. In addition, we add three Coord \cite{liu2018intriguing} layers to explicitly reinforce the positional information of the coarse mask. The layers consist of the values in range [-1,1] sorted by the height, the width, and the distance from the center. In conclusion, the specifier that we use in CRVOS has two channels for the previous frame's coarse mask and three channels for the coordinates information. We call our specifier the Clue, since it is simple but plays a key role.

\subsection{Refine Module} \label{refine module}
The general refine modules for Semi-VOS upscale the features with bilinear upsampling layers. However, a bilinear upsampling layer is not able to generate detailed spatial information since it is not trainable. Therefore, we use a deconvolution layer instead of a bilinear upsampling layer to generate more detailed spatial information. Since a deconvolution layer is usually time-consuming, we reduce the output channel size to 2 to avoid the speed decrement. Furthermore, we connect our refine modules with skip connections to fully exploit the features from each refine module. Our decoder can predict the current frame's mask more accurately by using multiple outputs from three refine modules. The mask is composed of two channels, one is the probabilities of the foreground, and the other the background.

\begin{table}[t]
	\centering
	\begin{tabular}{l|c c c c}
											 & O-FT   	  & fps 	      & $\mathcal{J}\uparrow$ & $\mathcal{F}\uparrow$ \\ \midrule
	MaskTrack \cite{perazzi2017learning} 	 & \checkmark & 0.08 		  & 79.7\% 				  & 75.4\%				  \\
	OSVOS \cite{caelles2017one} 			 & \checkmark & 0.11 		  & 79.8\% 				  & 80.6\% 				  \\
	OSVOS-S \cite{maninis2018video} 		 & \checkmark & \textbf{0.22} & 85.6\% 				  & \textbf{87.5}\% 	  \\
	OnAVOS \cite{voigtlaender2017online} 	 & \checkmark & 0.08 		  & \textbf{86.1}\%  	  & 84.9\%				  \\ \midrule
 	OSMN \cite{yang2018efficient} 			 & 			  & 7.14     	  & 74.0\%  			  & 72.9\%				  \\
	VideoMatch \cite{hu2018videomatch}  	 &   		  & 3.13   		  & 81.0\%  			  & -   				  \\
	FAVOS \cite{cheng2018fast} 			     & 			  & 0.56   		  & \textbf{82.4}\% 	  & 79.5\%  			  \\
	RGMP \cite{wug2018fast} 			     &    	      & 7.69   	      & 81.5\%   		      & 82.0\%  			  \\
	FEELVOS \cite{voigtlaender2019feelvos}   &  		  & 2.22 		  & 81.1\%   			  & \textbf{82.2}\%  	  \\
	A-GAME \cite{johnander2019generative}  	 &   		  & 14.3   	      & 81.5\%   	   	   	  & \textbf{82.2}\%  	  \\
	SiamMask \cite{wang2019fast}  	         &   	      & 55.0   	      & 71.7\%     	   	      & 67.8\%    	          \\
	\textbf{CRVOS(Ours)}    	  	   	     &     	      & \textbf{63.5} & 82.2\%     	          & 81.0\%                \\
	\end{tabular}
	\caption{Quantitative methods evaluated on DAVIS 2016 validation set by using fps, $\mathcal{J}$, and $\mathcal{F}$.}
	\label{results table 16}
\end{table}

\begin{table}[t]
	\centering
	\begin{tabular}{l | c c c c}
		                                     & O-FT       & fps           & $\mathcal{J}\uparrow$ & $\mathcal{F}\uparrow$ \\ \midrule
		OSVOS \cite{caelles2017one}          & \checkmark & 0.06          & 56.6\%                & 63.9\%                \\
		OSVOS-S \cite{maninis2018video}      & \checkmark & 0.11          & \textbf{64.7}\%       & \textbf{71.3}\%       \\
		OnAVOS \cite{voigtlaender2017online} & \checkmark & 0.04          & 64.5\%                & 71.2\%                \\ \midrule
		OSMN \cite{yang2018efficient}        &            & 3.57          & 52.5\%                & 57.1\%                \\
		FAVOS \cite{cheng2018fast}           &            & 0.28          & 54.6\%                & 61.8\%                \\
		RGMP \cite{wug2018fast}              &            & 3.85          & \textbf{64.8}\%       & \textbf{68.6}\%       \\
		SiamMask \cite{wang2019fast}         &            & \textbf{55.0} & 54.3\%                & 58.5\%                \\
		\textbf{CRVOS(Ours)}                 &            & 53.0          & 53.5\%                & 55.1\%
	\end{tabular}
	\caption{Quantitative methods evaluated on DAVIS 2017 validation set by using fps, $\mathcal{J}$, and $\mathcal{F}$.}
	\label{results table 17}
\end{table}

\subsection{Training Strategy} \label{training}
We use the two-step training strategy. First we pre-train the network on Youtube-VOS \cite{xu2018youtube_1} train set, and then fine-tune the network on DAVIS 2016 \cite{perazzi2016benchmark} train set. For pre-training and fine-tuning, we use randomly chosen 8 and 16 image frames as the input of the network respectively.

\subsection{Implementation Details}
We use negative log-likelihood (NLL) loss function and Adam optimizer. In pre-training stage, each input image is resized to $240\times432$, and the network is trained for 100 epochs with the learning rate of 1e-4. In fine-tuning stage, each input image has its original size, $480\times864$, and the network is trained for 500 epochs with the learning rate of 1e-5. To train the network with sufficient amount of data, we use horizontal filp, rotation($-$30$^\circ$$\sim$30$^\circ$), shearing ($-$30$^\circ$$\sim$30$^\circ$), and scaling (0.75$\sim$1.25) for data augmentation in fine-tuning stage. If there are multiple targets in the training process, one is chosen and dealt with at random. Whereas during the test, the masks of the entire targets are generated, and then overlapped.

\section{Experiments}
\subsection{Evaluation}
DAVIS 2016 \cite{perazzi2016benchmark} is the most popular dataset which only deals with single-target scenarios. In Table \ref{results table 16}, we show quantitative evaluations on DAVIS 2016 validation set. Our proposed method is the most efficient method with 63.5 fps and $\mathcal{J}\&\mathcal{F}$ score of 81.6\%. DAVIS 2017 \cite{pont20172017} is an extended version of DAVIS 2016, also including multi-target scenarios. In Table \ref{results table 17}, we show the evaluations on DAVIS 2017 validation set. Our method shows 53.0 fps and $\mathcal{J}\&\mathcal{F}$ score of 54.3\%. The speed of our proposed method is calculated on a single GeForce RTX 2080 Ti GPU.

\subsection{Ablation Studies} \label{ablation}
\begin{table}[t]
	\centering
	\begin{tabular}{l c c c | c c c}
		& RM & PM & Clue & $\mathcal{J}\&\mathcal{F}\uparrow$ & $\mathcal{J}\uparrow$ & $\mathcal{F}\uparrow$\\
		\midrule
		\Romannum{1} & \checkmark & & & 72.1\% & 72.3\% & 71.9\%\\
		\Romannum{2} & \checkmark & \checkmark & & 78.6\% & 78.7\% & 78.4\%\\
		\Romannum{3} & \checkmark & & \checkmark & \textbf{81.6}\% & \textbf{82.2}\% & \textbf{81.0}\%\\
		\Romannum{4} & & & \checkmark & 79.1\% & 79.3\% & 78.8\%\\
	\end{tabular}
	\caption{Ablation studies on DAVIS 2016 validation set. RM indicates using our novel refine modules instead of general ones. PM indicates using previous frame's coarse mask as a specifier, and Clue indicates using the Clue as a specifier.}
	\label{ablation table}
\end{table}

We conduct ablation studies to demonstrate our work. In Table \ref{ablation table}, we compare the accuracy of the four networks with different specifiers and refine modules.

\vspace{1mm}
\noindent\textbf{Specifier:} We show the effect of using different specifiers by comparing \Romannum{1}, \Romannum{2} and \Romannum{3}. \Romannum{1} is the network with no specifier. Since it is hard to specify the given target, it shows the worst accuracy, $\mathcal{J}\&\mathcal{F}$ score of 72.1\%. \Romannum{2} uses the previous frame's coarse mask as a specifier. It shows $\mathcal{J}\&\mathcal{F}$ score of 78.6\% which is a lot better than \Romannum{1}. \Romannum{3} uses the Clue, i.e. the previous frame's coarse mask with coordinates information, as a specifier. Since coordinates information can reinforce positional information of the previous frame's coarse mask, \Romannum{3} shows the best accuracy, $\mathcal{J}\&\mathcal{F}$ score of 81.6\%.

\begin{figure*}[h]
	\centering
	\includegraphics[width=1.0\linewidth]{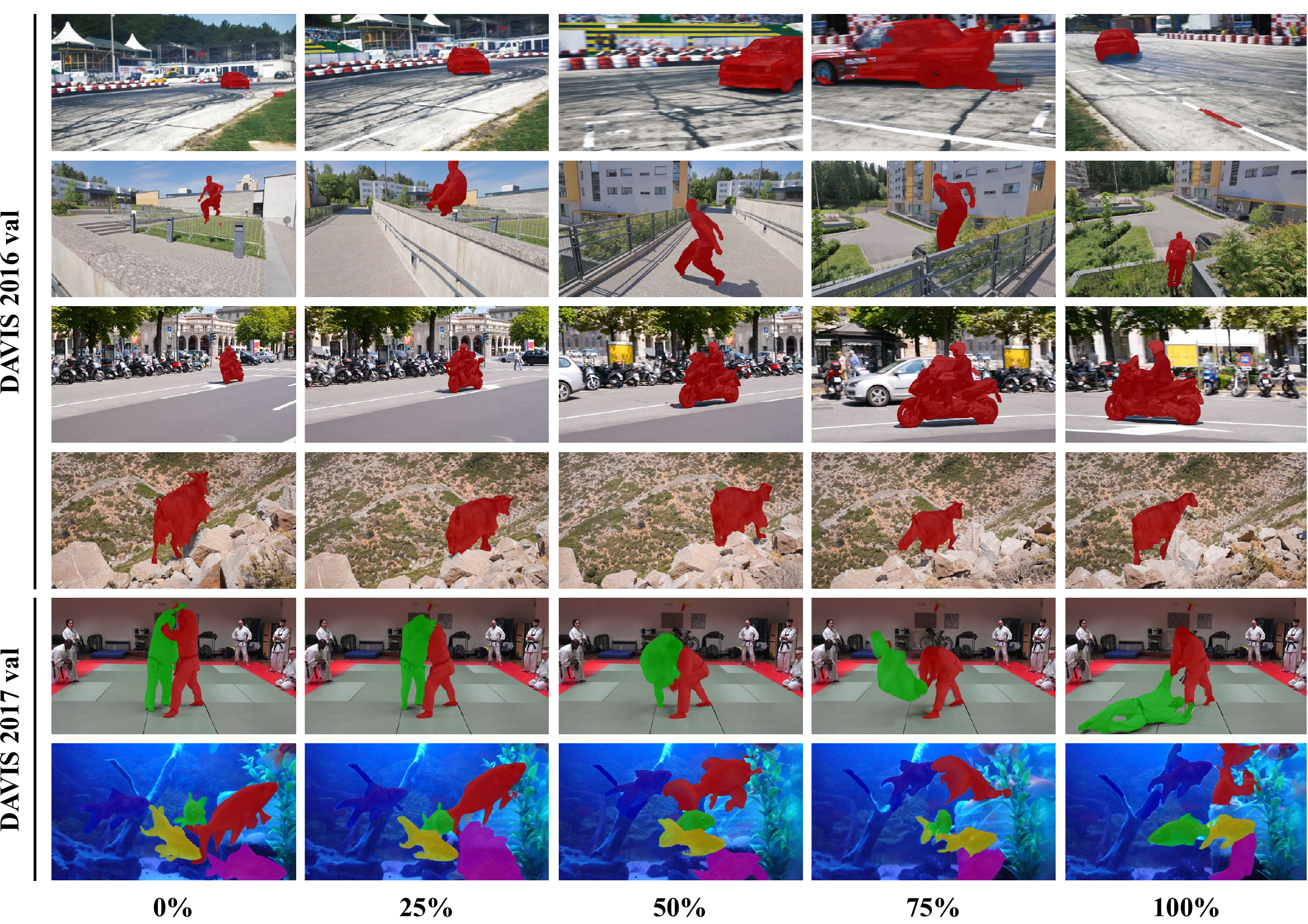}
	\caption{Qualitative results of CRVOS on DAVIS 2016 validation set and DAVIS 2017 validation set. The numbers below the figure indicate the ratio of the past frames to the total frames. 0\% indicates the initial frame and 100\% indicates the last frame. As seen in the figure, CRVOS consistently produces high quality segmentations in relatively simple scenarios.}
	\label{figure3}
\end{figure*}

\vspace{1mm}
\noindent\textbf{Refine module:} We show the effect of using different refine modules by comparing \Romannum{3} and \Romannum{4}. \Romannum{3} uses our refine module that is composed of a convolution layer, a deconvolution layer, and a skip connection that connects three refine modules. On the other hand, \Romannum{4} uses a general refine module instead of our refine module. A deconvolution layer is replaced by a convolution layer that reduces the feature dimension, and a bilinear upsampling layer that increases the feature size. Also, there is no skip connection that connects multiple refine modules. It can be seen that \Romannum{4} shows the lower accuracy compared to \Romannum{3}, $\mathcal{J}\&\mathcal{F}$ score of 79.1\%.

\subsection{Qualitative Results} \label{qualitative results}
Fig. \ref{figure3} shows the qualitative results of CRVOS. The upper four rows are the results on DAVIS 2016 \cite{perazzi2016benchmark} validation set which deals with single-target scenarios, and the rest are the results on DAVIS 2017 \cite{pont20172017} validation set which deals with multi-target scenarios.

\vspace{1mm}
\noindent\textbf{DAVIS 2016:} In the first scenario, the target moves fast and the scale of the target rapidly changes. In this case, the network generates poor quality segmentations since the Clue provides inaccurate information about the given target. The second, the third, and the fourth scenarios are comparatively simple scenarios, because they have single targets with small movements. It can be observed that the network can produce precise segmentations in theses cases.

\vspace{1mm}
\noindent\textbf{DAVIS 2017:} In the fifth and the last scenarios, there are multiple targets with small movements. The results show that if positional changes of the targets between adjacent frames are not large, the network is able to generate the accurate target masks in multi-target scenarios as well.

\section{Conclusion}
In this work, we have proposed a novel real-time network for Semi-VOS. By using the Clue as a specifier without any intermediate network, our proposed method achieves the fastest speed among the existing methods. The results on the benchmark datasets also demonstrate that our method shows comparable accuracy to the state-of-the-art methods when dealing with relatively simple scenarios. 

\vspace*{\fill}
\noindent\footnotesize\textbf{Acknowledgement. }This research was supported by Multi-Ministry Collaborative R\&D Program(R\&D program for complex cognitive technology) through the National Research Foundation of Korea(NRF) funded by MSIT, MOTIE, KNPA(NRF-2018M3E3A1057289).

\clearpage
\bibliographystyle{IEEEbib}
\small \bibliography{strings,refs}

\begin{thebibliography}{10}

\bibitem{perazzi2016benchmark}
Federico Perazzi, Jordi Pont-Tuset, Brian McWilliams, Luc Van~Gool, Markus
  Gross, and Alexander Sorkine-Hornung,
\newblock ``A benchmark dataset and evaluation methodology for video object
  segmentation,''
\newblock in {\em Proceedings of the IEEE Conference on Computer Vision and
  Pattern Recognition}, 2016, pp. 724--732.

\bibitem{caelles2017one}
Sergi Caelles, Kevis-Kokitsi Maninis, Jordi Pont-Tuset, Laura Leal-Taix{\'e},
  Daniel Cremers, and Luc Van~Gool,
\newblock ``One-shot video object segmentation,''
\newblock in {\em Proceedings of the IEEE conference on computer vision and
  pattern recognition}, 2017, pp. 221--230.

\bibitem{voigtlaender2017online}
Paul Voigtlaender and Bastian Leibe,
\newblock ``Online adaptation of convolutional neural networks for video object
  segmentation,''
\newblock {\em arXiv preprint arXiv:1706.09364}, 2017.

\bibitem{perazzi2017learning}
Federico Perazzi, Anna Khoreva, Rodrigo Benenson, Bernt Schiele, and Alexander
  Sorkine-Hornung,
\newblock ``Learning video object segmentation from static images,''
\newblock in {\em Proceedings of the IEEE Conference on Computer Vision and
  Pattern Recognition}, 2017, pp. 2663--2672.

\bibitem{hu2017maskrnn}
Yuan-Ting Hu, Jia-Bin Huang, and Alexander Schwing,
\newblock ``Maskrnn: Instance level video object segmentation,''
\newblock in {\em Advances in Neural Information Processing Systems}, 2017, pp.
  325--334.

\bibitem{marki2016bilateral}
Nicolas M{\"a}rki, Federico Perazzi, Oliver Wang, and Alexander
  Sorkine-Hornung,
\newblock ``Bilateral space video segmentation,''
\newblock in {\em Proceedings of the IEEE Conference on Computer Vision and
  Pattern Recognition}, 2016, pp. 743--751.

\bibitem{jampani2017video}
Varun Jampani, Raghudeep Gadde, and Peter~V Gehler,
\newblock ``Video propagation networks,''
\newblock in {\em Proceedings of the IEEE Conference on Computer Vision and
  Pattern Recognition}, 2017, pp. 451--461.

\bibitem{cheng2018fast}
Jingchun Cheng, Yi-Hsuan Tsai, Wei-Chih Hung, Shengjin Wang, and Ming-Hsuan
  Yang,
\newblock ``Fast and accurate online video object segmentation via tracking
  parts,''
\newblock in {\em Proceedings of the IEEE Conference on Computer Vision and
  Pattern Recognition}, 2018, pp. 7415--7424.

\bibitem{yang2018efficient}
Linjie Yang, Yanran Wang, Xuehan Xiong, Jianchao Yang, and Aggelos~K
  Katsaggelos,
\newblock ``Efficient video object segmentation via network modulation,''
\newblock in {\em Proceedings of the IEEE Conference on Computer Vision and
  Pattern Recognition}, 2018, pp. 6499--6507.

\bibitem{wug2018fast}
Seoung Wug~Oh, Joon-Young Lee, Kalyan Sunkavalli, and Seon Joo~Kim,
\newblock ``Fast video object segmentation by reference-guided mask
  propagation,''
\newblock in {\em Proceedings of the IEEE Conference on Computer Vision and
  Pattern Recognition}, 2018, pp. 7376--7385.

\bibitem{johnander2019generative}
Joakim Johnander, Martin Danelljan, Emil Brissman, Fahad~Shahbaz Khan, and
  Michael Felsberg,
\newblock ``A generative appearance model for end-to-end video object
  segmentation,''
\newblock in {\em Proceedings of the IEEE Conference on Computer Vision and
  Pattern Recognition}, 2019, pp. 8953--8962.

\bibitem{lin2019agss}
Huaijia Lin, Xiaojuan Qi, and Jiaya Jia,
\newblock ``Agss-vos: Attention guided single-shot video object segmentation,''
\newblock in {\em Proceedings of the IEEE International Conference on Computer
  Vision}, 2019, pp. 3949--3957.

\bibitem{zeng2019dmm}
Xiaohui Zeng, Renjie Liao, Li~Gu, Yuwen Xiong, Sanja Fidler, and Raquel
  Urtasun,
\newblock ``Dmm-net: Differentiable mask-matching network for video object
  segmentation,''
\newblock in {\em Proceedings of the IEEE International Conference on Computer
  Vision}, 2019, pp. 3929--3938.

\bibitem{oh2019video}
Seoung~Wug Oh, Joon-Young Lee, Ning Xu, and Seon~Joo Kim,
\newblock ``Video object segmentation using space-time memory networks,''
\newblock {\em arXiv preprint arXiv:1904.00607}, 2019.

\bibitem{wang2019ranet}
Ziqin Wang, Jun Xu, Li~Liu, Fan Zhu, and Ling Shao,
\newblock ``Ranet: Ranking attention network for fast video object
  segmentation,''
\newblock in {\em Proceedings of the IEEE International Conference on Computer
  Vision}, 2019, pp. 3978--3987.

\bibitem{he2016deep}
Kaiming He, Xiangyu Zhang, Shaoqing Ren, and Jian Sun,
\newblock ``Deep residual learning for image recognition,''
\newblock in {\em Proceedings of the IEEE conference on computer vision and
  pattern recognition}, 2016, pp. 770--778.

\bibitem{deng2009imagenet}
Jia Deng, Wei Dong, Richard Socher, Li-Jia Li, Kai Li, and Li~Fei-Fei,
\newblock ``Imagenet: A large-scale hierarchical image database,''
\newblock in {\em 2009 IEEE conference on computer vision and pattern
  recognition}. Ieee, 2009, pp. 248--255.

\bibitem{liu2018intriguing}
Rosanne Liu, Joel Lehman, Piero Molino, Felipe~Petroski Such, Eric Frank, Alex
  Sergeev, and Jason Yosinski,
\newblock ``An intriguing failing of convolutional neural networks and the
  coordconv solution,''
\newblock in {\em Advances in Neural Information Processing Systems}, 2018, pp.
  9605--9616.

\bibitem{maninis2018video}
K-K Maninis, Sergi Caelles, Yuhua Chen, Jordi Pont-Tuset, Laura Leal-Taix{\'e},
  Daniel Cremers, and Luc Van~Gool,
\newblock ``Video object segmentation without temporal information,''
\newblock {\em IEEE transactions on pattern analysis and machine intelligence},
  vol. 41, no. 6, pp. 1515--1530, 2018.

\bibitem{hu2018videomatch}
Yuan-Ting Hu, Jia-Bin Huang, and Alexander~G Schwing,
\newblock ``Videomatch: Matching based video object segmentation,''
\newblock in {\em Proceedings of the European Conference on Computer Vision
  (ECCV)}, 2018, pp. 54--70.

\bibitem{voigtlaender2019feelvos}
Paul Voigtlaender, Yuning Chai, Florian Schroff, Hartwig Adam, Bastian Leibe,
  and Liang-Chieh Chen,
\newblock ``Feelvos: Fast end-to-end embedding learning for video object
  segmentation,''
\newblock in {\em Proceedings of the IEEE Conference on Computer Vision and
  Pattern Recognition}, 2019, pp. 9481--9490.

\bibitem{wang2019fast}
Qiang Wang, Li~Zhang, Luca Bertinetto, Weiming Hu, and Philip~HS Torr,
\newblock ``Fast online object tracking and segmentation: A unifying
  approach,''
\newblock in {\em Proceedings of the IEEE Conference on Computer Vision and
  Pattern Recognition}, 2019, pp. 1328--1338.

\bibitem{xu2018youtube_1}
Ning Xu, Linjie Yang, Yuchen Fan, Dingcheng Yue, Yuchen Liang, Jianchao Yang,
  and Thomas Huang,
\newblock ``Youtube-vos: A large-scale video object segmentation benchmark,''
\newblock {\em arXiv preprint arXiv:1809.03327}, 2018.

\bibitem{pont20172017}
Jordi Pont-Tuset, Federico Perazzi, Sergi Caelles, Pablo Arbel{\'a}ez, Alex
  Sorkine-Hornung, and Luc Van~Gool,
\newblock ``The 2017 davis challenge on video object segmentation,''
\newblock {\em arXiv preprint arXiv:1704.00675}, 2017.

\end{thebibliography}
\end{document}